\documentclass[conference]{IEEEtran}
\IEEEoverridecommandlockouts
\usepackage{cite}
\usepackage{amsmath,amssymb,amsfonts}
\usepackage{algorithmic}
\usepackage{graphicx}
\usepackage{textcomp}
\usepackage{xcolor}
\usepackage{array}%需要用到该宏包
\usepackage{footnote}
\makesavenoteenv{tabular}

\def\BibTeX{{\rm B\kern-.05em{\sc i\kern-.025em b}\kern-.08em
    T\kern-.1667em\lower.7ex\hbox{E}\kern-.125emX}}
\begin{document}

\title{LLMCount: Enhancing Stationary mmWave Detection with Multimodal-LLM}

\author{\IEEEauthorblockN{Boyan Li}
\IEEEauthorblockA{
\textit{The Chinese Univeristy of Hong Kong, Shenzhen}\\
% Shenzhen, China \\
119010142@link.cuhk.edu.cn}
\and
\IEEEauthorblockN{Shengyi Ding}
\IEEEauthorblockA{
\textit{Hong Kong University of Science and Technology(Guangzhou)}\\
% Beijing, China \\
dsy2022140367@bupt.edu.cn}
\and
\IEEEauthorblockN{Deen Ma}
\IEEEauthorblockA{
\textit{Shenzhen University}\\
% Chengdu, China \\
2210433107@email.szu.edu.cn}
\and
\IEEEauthorblockN{Yixuan Wu}
\IEEEauthorblockA{
\textit{Zhejiang University}\\
% Hangzhou, China \\
wyx\_chloe@zju.edu.cn}
\and
\IEEEauthorblockN{Hongjie Liao}
\IEEEauthorblockA{
\textit{The Chinese University of Hong Kong, Shenzhen}\\
% Shenzhen, China \\
lhj984222431@gmail.com}
\and
\IEEEauthorblockN{~~~~~~~~~~~~~~~~~~~~~~~~~~~~~~~~~~~~~~~~~~~Kaiyuan Hu*}
\IEEEauthorblockA{
\textit{~~~~~~~~~~~~~~~~~~~~~~~~~~~~~~~~~~~~~~~~~~~~~The Chinese University of Hong Kong, Shenzhen}\\
% Shenzhen, China\\
~~~~~~~~~~~~~~~~~~~~~~~~~~~~~~~~~~~~~~~~~~~~~222010505@link.cuhk.edu.cn}
}

\maketitle

\begin{abstract}
Millimeter wave sensing provides people with the capability of sensing the surrounding crowds in a non-invasive and privacy-preserving manner, which holds huge application potential. However, detecting stationary crowds remains challenging due to several factors such as minimal movements (like breathing or casual fidgets), which can be easily treated as noise clusters during data collection and consequently filtered in the following processing procedures. Additionally, the uneven distribution of signal power due to signal power attenuation and interferences resulting from external reflectors or absorbers further complicates accurate detection. To address these challenges and enable stationary crowd detection across various application scenarios requiring specialized domain adaption, we introduce LLMCount, the first system to harness the capabilities of large-language models (LLMs) to enhance crowd detection performance. By exploiting the decision-making capability of LLM, we can successfully compensate the signal power to acquire a uniform distribution and thereby achieve a detection with higher accuracy. To assess the system's performance, comprehensive evaluations are conducted under diversified scenarios like hall, meeting room, and cinema. The evaluation results show that our proposed approach reaches high detection accuracy with lower overall latency compared with previous methods.
\end{abstract}

\vspace{-2.5mm}
\section{Introduction}
\label{sec: intro}
%background
Millimeter wave (mmWave) sensing enables a natural way to detect surrounding objects, which has shown huge potential in various application scenarios like crowd detection, transportation management\cite{8254900}, or industrial automation\cite{7032050}, with its non-invasive and privacy-preserving features. Among the various application scenarios, crowd detection shows huge potential in providing necessary information for healthcare \cite{10.1145/2942358.2942381}, education, or entertainment services. Traditional methods like camera-based approaches \cite{yang2021vision} utilize computer vision detection algorithms to detect the crowd from captured images on a specific position. However, such approaches raise the problem of privacy disclosure and are often affected by limited illumination conditions. To avoid such defects, WiFi-based approaches \cite{wu2015non} are proposed. Whilst, such approaches ignore the detection accuracy, lacking the ability to identify the precise location of each individual in the crowd. To tackle the above-mentioned problems, mmWave sensing approaches \cite{cheng2021robust} are adopted. In addition, as the main medium of emerging 5G communications, mmWave also contains the potential to be exploited as a ubiquitous sensing tool in the upcoming artificial general intelligence era.

%describe how similar previous works performed, in order to introduce the challenges we are currently facing: 1. inadequate parameter adjustment for data enhancement process 2. high computation power consumption on local devices 3. previous method lack the ability to generalize to different sensors (we can not use the same parameters setups for all kinds of situations or all kinds of sensors we use)
Previous works of mmWave sensing for crowd detection have shown their capability of sensing the crowd at satisfactory accuracy. However, we still face some challenges:
Firstly, parameter optimization in the data enhancement process often lacks precision, leading to either data distortion or over-enhancement. This misadjustment significantly affects the accuracy and reliability of target detection, undermining the overall system performance \cite{amirgholipour2021pdanet}.
Secondly, the high computational demands of mmWave systems strain local device resources, increasing operational costs and limiting deployment in energy-constrained environments. Efficient algorithmic improvements are essential to balance computational intensity with power consumption without sacrificing detection accuracy \cite{choi2022wi}.
Lastly, the ability of mmWave sensing to generalize across different sensors is limited, as existing methods are often inflexible to varying operational conditions and sensor characteristics. Enhancing adaptability is crucial for scaling the technology across diverse deployment scenarios and sensor types \cite{gao2024comprehensive}.

%以上的三个challenge对应我们的三个contribution
To tackle the above-mentioned challenges, we hereby propose LLMCount, aiming to exploit the power of multimodal large-language models (MLLM) to enhance the sensing performance of millimeter-wave sensors. We conclude our contributions as follows:
%the order of contribution 2&3 
\begin{itemize}
    \item We achieved higher detection performance by harnessing the decision-making power of MLLM to gain better data processing capability.
    \item By offloading most processing to the cloud, we mitigate the local computation consumption and increase the detection frequency.
    \item The system's ability to generalize across diverse devices and scenarios in detection is significantly enhanced.
\end{itemize}
\begin{figure}[t]
\centering
\includegraphics[width=\columnwidth]{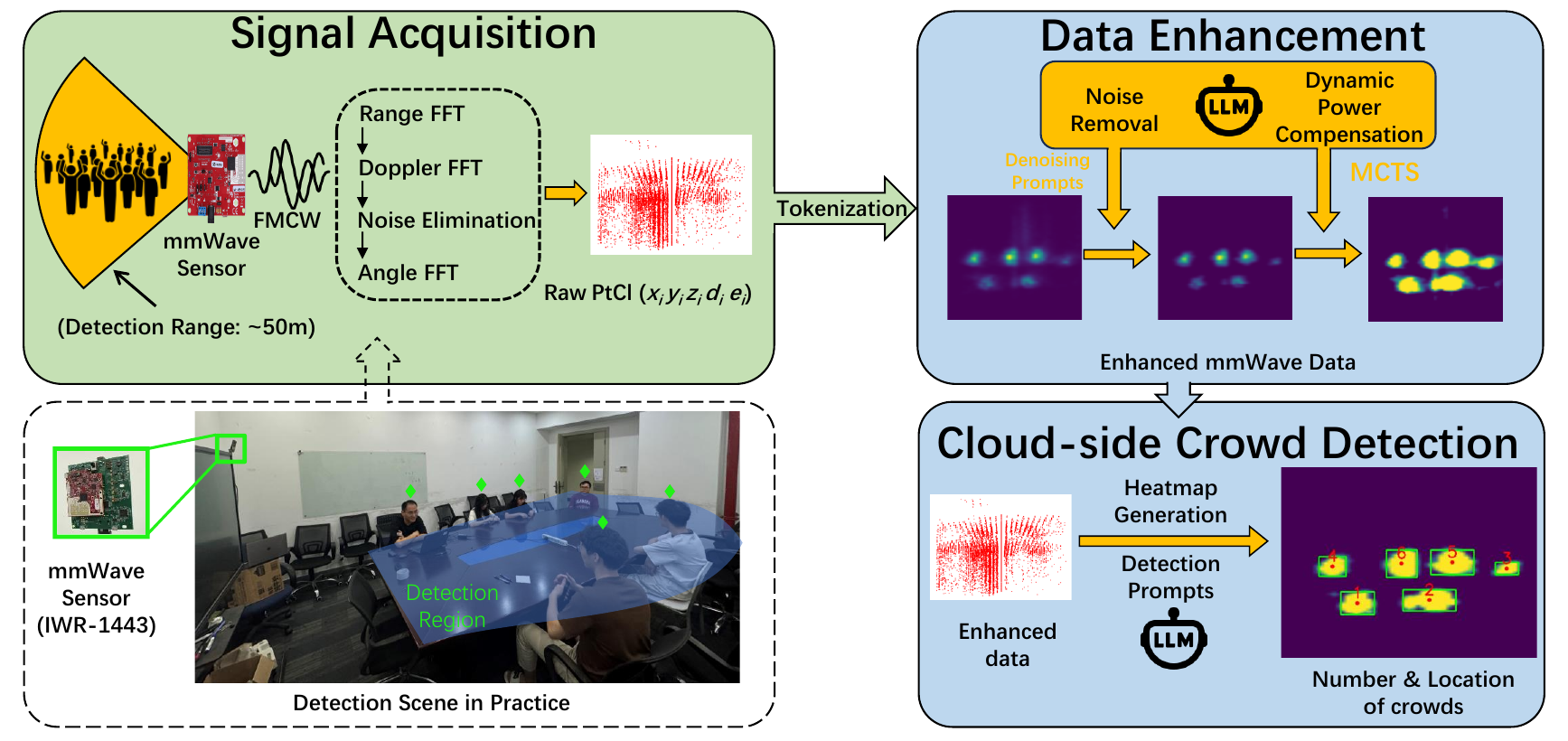}
 \caption{System Architecture and workflow of LLMCount\\Note: To facilitate observation \& comparison, we visualize the intermediate data in this workflow as heatmaps.}
 \label{framework}
\end{figure}
\vspace{-2.5mm}
\section{System Design}
% \subsection{System Overview}
% \label{sec: system overview}
We illustrate the system architecture of LLMCount in Figure \ref{framework}, which consists of three modules, with one running on local processing units (shown in light green) and two deployed on multimodal-LLM as processing agents (shown in light blue). By leveraging LLMCount, users can collect the raw mmWave data using commodity radar chips and upload the raw data to the cloud-side agents for detection and get the detection results in real time. The working region for the system covers a region with a radius of about 50 meters (varies depending on sensor type) which is capable of most application scenarios.
\subsection{Signal Acquisition}
%Collection of the mmWave data using commodity mmWave radar chips: 
In this work, we use IWR-1443 by Texas Instrument as the testbed, the working region covers an area of half-round with a radius of 50 meters. The signal acquisition module is composed of a mmWave radar chip deployed on an edge computing unit with limited computation resources (Nvidia Jetson Nano). The collected data is initially preprocessed locally and output in the format of raw point clouds with noise interference included. Then, the data is output as coordinates and instantaneous velocities of the points generated by the radar. In practice, during each short interval referred to as a 'frame', the radar generates multiple points. Each of these points is represented by a vector: 
\begin{equation}
\label{eq: data_format}
    p_{i} = (x_{i},y_{i},z_{i},d_{i},e_{i}) 
\end{equation} in which $i$ refers to the index of this point in current frame, $x_{i}$,$y_{i}$ and $z_{i}$ refers to the point's spatial coordinates (relative position to the radar antenna), $d_{i}$ refers to doppler, the relative velocity to the radar antenna, the $e_{i}$ refers to the energy intensity of the reflected signal. 

\begin{figure}[t]
    \centering
    \includegraphics[width=\linewidth]{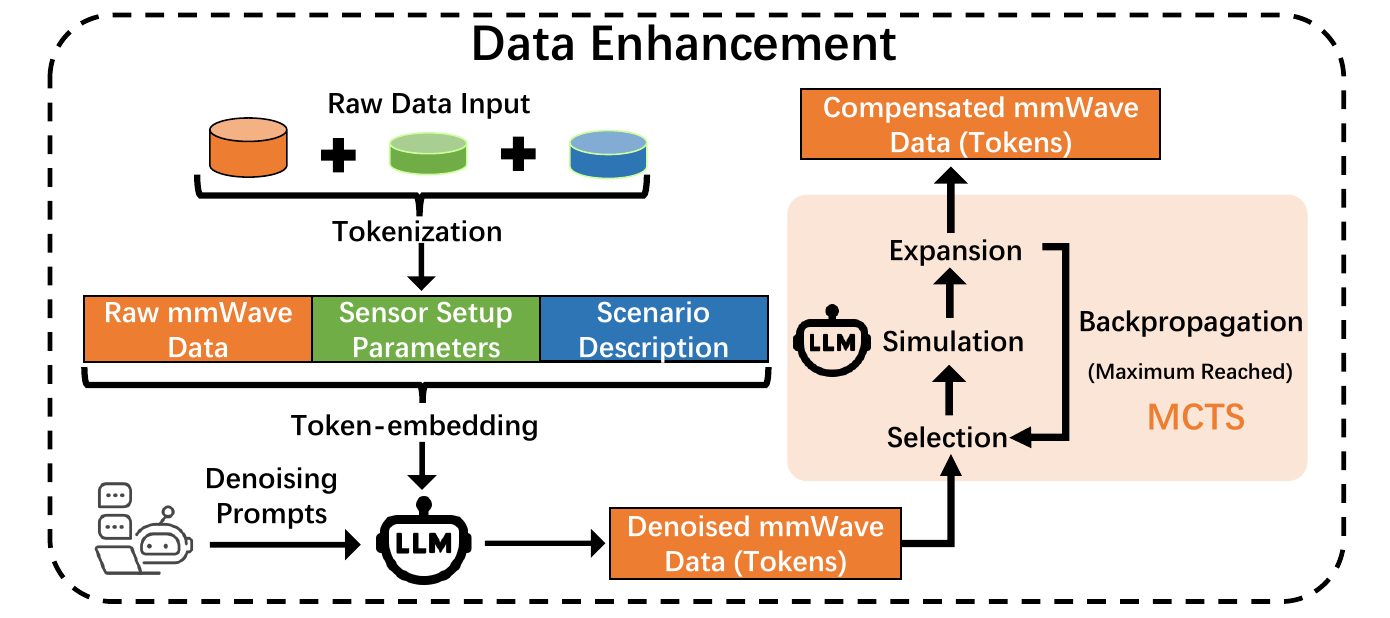}
    \caption{Data Enhancement Worklow}
    \label{fig: Data Enhancement}
\end{figure}

\subsection{Data Enhancement}
%这部分得专门画个workflow图-DSY
%citation of multipath effect and environmental reflection-MDE

Collected raw mmWave data are often accompanied by noise and distributed ununiformly as results of multipath effect \cite{gustafson2013mm}, environmental reflection \cite{zhang2019reflections}, and propagation attenuation \cite{norouzian2019rain} which can not be directly employed for crowd detection. To prepare the raw data for accurate detection, we utilize the decision-making power of MLLM to enhance the raw data. In this work, we transform the noise removal process along with signal power compensation to MLLM to leverage its adaptive capability. The overall workflow is demonstrated in Figure \ref{fig: Data Enhancement}.

\noindent\textbf{Noise Removal}\\
In this part, we focus on optimizing the noise removal process by harnessing the generalization and analyzing capability of MLLM. We first convert the raw data into tokens as part of the fundamental input. To further enhance the noise removal performance, we provide the setup information of the mmWave sensors as auxiliary data to provide contextual interpretation of the raw data, including frequency range, chirp configuration, plus range and doppler resolutions, which help to allocate different processing granularity. These features are often neglected in previous works \cite{zhang2022synthesized}. Since interference factors vary depending on the working scenarios, we include a description of the current working scenario to guide more informed decisions about what may constitute noise versus valid signal data. To ensure a standardized input for context and scenario descriptions, we define six attributes to describe current working scenarios: \textit{Environment Type (Indoor/Outdoor), Surface Characteristics (Smooth/Rough), Material Properties (Metallic/Non-Metallic), Crowd Density (Sparse/Dense), Motion Dynamic (Static/Dynamic), and Obstacles (Obstructed/Unobstructed)}. Such an approach ensures a more stable and efficient processing. The final step is to direct the MLLM's focus by providing specifically crafted prompts to effectively utilize the single input sequence for the noise removal process.

In practice, we integrate and encode the data of four modalities into token-based embeddings, enabling effective processing by the MLLM. Through tokenization, we convert the original inputs into a unified sequence of tokens. This approach allows the MLLM to analyze and process the data holistically, enhancing noise removal performance. To achieve efficient processing for real-time detection, we fix the length of the token sequences for the raw data by allocating a fixed window for the acquired mmWave data of 200ms. As shown in Figure \ref{fig: Data Enhancement}, the four input modalities are converted into tokens and then concatenated into a single sequence. 
% The results are illustrated in Figure. \ref{},  %DSY画图
% which demonstrates the visualization of the data before and after the denoising process.\\

%这里再加一个noise removal前后的图

% \noindent\textbf{LLM-assisted dynamic power compensation}\\

\noindent\textbf{Dynamic Power Compensation}\\
Previous research has shown the signal power attenuation effect \cite{wang2021millimeter} for both emitted and received signals during propagation \cite{mehrotra2019wave}, which can be represented as the following formula:
\[P_L=L_d+L_{Rl}=20\log\left(\frac{4\pi d}{\lambda}\right)+L_{Rl}\]
Where, Ld and LRl represent the signal attenuation due to distance and signal reflection loss, respectively. We ignore the LRl here as it is not relevant to the energy loss in the scenario under discussion.

%formula参考ICASSP2024，如果有更准确的，可以替换成新的
While existing methods \cite{mmCount} utilize the formula corresponding to the radar chip configurations to fit the trend of signal attenuation and compensate for it, these methods often fail to generalize across different scenarios where radar parameters and working scenarios vary. In this step, we inherit the input of the setup description and scenarios description in the previous step to assist power compensation, ensuring more robust and adaptable performance across diverse conditions. 

To extend beyond the limitations of the traditional method in handling complex data compensation tasks, we propose to integrate LLM with Monte Carlo Tree Search (MCTS). Based on pre-defined prompts, LLM generates various compensation strategies, while MCTS searches and optimizes these strategies, identifying the most effective compensation path. The state of MCTS represents the current condition of the data being compensated. Action space in this workflow is selecting a specific prompt, which guides the system in further modifying or amplifying the data, such as Prompts to adjust specific data features or prompts that guide LLM. The main steps in the workflow of MCTS are organized as follows:

\begin{itemize}
\item \textbf{Initialization: } Prior to executing the MCTS algorithm, we initialize by selecting raw data as the starting state and constructing a root node incorporating this raw data.
\item \textbf{Selection: } MCTS algorithm navigates through the tree from the root using a pre-defined strategy. The Upper Confidence Bound (UCB) formula is employed to balance exploration and exploitation when selecting the most promising child node (action), described as follows:
\[UCB(s_i)=\hat{W}(s_i)+c\sqrt{\frac{lnN}{n(s_i)}}\]
where \(\hat{W}(s_i)\) is the average score of the compensation in node \(s_i\). \(N\) is total number of visits across all nodes up to the current moment. \(n(s_i)\) is the number of visits to node \(s_i\). \(c\) is a constant that balances exploration versus exploitation. 

\item \textbf{Simulation: }After selecting an action, the algorithm proceeds with the simulation phase, where the LLM generates a compensation result and calculates the compensation score.

 The compensation score \(W\) can be calculated by
\[W = \mu_1 E + \mu_2 A + \mu_3 S\]

where \(E\) represents compensation Effectiveness, which evaluates the overall quality improvement after the LLM-based compensation, \(A\) represents compensation Accuracy, focuses on how closely the compensation data meets certain predefined targets, \(S\) is compensation Stability, measuring how consistently the compensation data performs across multiple iterations or variations in prompt usage. 
% where \(E\) represents compensation Effectiveness, which evaluates the overall quality improvement after the LLM-based compensation. It can be calcuted by the relative change of density in certain areas:
% \[E=\frac{Density(Compensation) - Density(Original)} {Density(Original)}\]
% \(A\) represents compensation Accuracy, focuses on how closely the compensation data meets certain predefined targets. It can be calculated by reducing RMSE or improving key features in the dataset:
% \[A=1-\frac{\sqrt{\frac{1}{n}\sum^n_{i=1}(D_{i,Compensation}-D_{i,Expected})^2}}{R_{max}}\]
% where \(D\) is data density, \(R_{max}\) is a parameter that larger than the maximum aloowable RMSE, ensuring the score ranges from 0 to 1.

% \(S\) represents compensation Stability, measuring how consistently the compensation data performs across multiple iterations or variations in prompt usage. It can be calculated by the variance of Density change:
% \[S=1 - \frac{\sigma^2(Density(Compensation))}{\sigma^2(Density(Original))}\]
\item \textbf{Expansion:} Once a local prompt is selected and executed, the expansion phase generates a new compensation scheme for the partial data. Subsequently, a new prompt is generated to dynamically adjust the data, which is then incorporated into the tree as a new action along with the corresponding new state. 

\item \textbf{Backpropagation:} After the simulation, backpropagation is the process of propagating the results of a simulation from the leaf nodes back to the root node. For each node sisi, the visit count \(n(s_i)\) and total compensation score \(W(s_i)\) are updated based on the current simulation. Specifically, the visit count is incremented by one and the total compensation score is increased by the compensation score \(W\) obtained from the simulation.
\[n(s_i) = n(s_i) + 1, W(s_i)=W(s_i)+W \]
After updating the visit count and total compensation score, the average compensation score \(\hat{W}(s_i)\) for node \(s_i\) is calculated as:
\[\hat{W}(s_i)=\frac{W(s_i)}{n(s_i)} \]
\item \textbf{Termination:} The algorithm repeats these phases until termination conditions are met, such as early stopping (when the improvement in search results diminishes or when repeated results are generated), search constraints (when the predetermined number of rollouts is reached or when one or more nodes in the tree reach the maximum depth constraint), and criteria based on LLM logits.
\end{itemize}
We compare the dynamic power compensation results of the traditional method \cite{mmCount} with our proposed LLM-based approach to demonstrate its effectiveness. As shown in Figure \ref{Traditional Method Detection} and Figure \ref{LLM-Assisted Detection}, the LLM-based method provides more effective compensation at corner positions, ensuring higher robustness in detection.

%下面这个图是不是上部分不全呀？
% 这两张图就是这样，在最上方有一个物体 --Boyan
%got it~ --Kaiyuan 
\iffalse
\begin{figure}[htbp]
    \centering
    \begin{minipage}[t]{0.45\columnwidth}
        \centering
        \includegraphics[width=\textwidth]{Figures/Traditional_Method.png}
        \caption{Traditional Compensation}
        \label{Traditional Method Detection}
    \end{minipage}
    \hfill
    \begin{minipage}[t]{0.45\columnwidth}
        \centering
        \includegraphics[width=\textwidth]{Figures/Raw_data_to_LLM.png}
        \caption{LLM-based Compensation}
        \label{LLM-Assisted Detection}
    \end{minipage}
\end{figure}
\fi

% Compare the results between LLM-assisted and traditional methods\\

\subsection{Cloud-side Crowd Detection}
% idea:how to distinguish real clusters and ghost clusters by MLLM?
\noindent\textbf{Heatmap Generation}\\
Once the signal is enhanced, the data is converted into heatmaps based on the x-y plane and intensity levels as described in Equation \ref{eq: data_format}. To incorporate intensity factors into the heatmap, each point \(p_i\) is repeated according to its intensity value. For example, a point at\((1,0)\) with an intensity of 6 will be replicated six times in the heatmap generation.
Next, we segment the data into sub-regions \(S_j\)  and count the number of points \(P_i\) in each segment:
\[P_j=\sum_{(x_i, y_i)\in S_j}n(x_i, y_i)\]
where \(n(x_i, y_i)\) represents there are \(n\) points  in \(x_i, y_i\).

% After adjusting for any loss of data using an compensation algorithm, we generate a false-color heatmap where the colors correspond to the number of points in each sub-region. We also create another heatmap with a controlled color range, where the intensity of the colors  \( C(x, y)\) represents the density of points.
Then we generate a heatmap with a controlled color range, the intensity of the colors \( C(x, y)\) represents the density of points.

\[C(x,y)=\frac{\hat{P_j}}{A_j}\]
where \(A_j\) is the area of sub-region \(S_j\) and \(\hat{P_j}\) is the number of points in sub-region \(S_j \)

\noindent\textbf{LLM-assisted Detection}
The final step of LLMCount is to utilize LLM to detect and annotate each individual in the crowd. %Leveraging the LLM's advanced semantic understanding and contextual reasoning capabilities, the model can accurately identify and annotate the position and number of each valid point cloud from video or image data, allowing for efficient automated data processing. 
To minimize the detection performance fluctuations, we implement a sliding time window strategy and set overlap between consecutive windows, ensuring consistent and accurate counting even in dynamic environments. This approach significantly enhances the accuracy and reliability of valid point cloud counting, as demonstrated in the experimental results.
% \begin{figure}[htbp]
%     \centering
%     \includegraphics[width=0.5\linewidth]{Figures/five_chopped_noted.png}
%     \caption{Cloud Detection}
%     \label{fig:enter-label}
% \end{figure}
\vspace{-2.5mm}
\begin{figure}[h]
    \centering
    \begin{minipage}[t]{0.30\columnwidth}
        \centering
        \includegraphics[width=\textwidth]{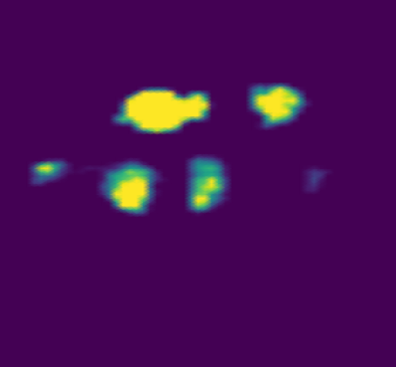}
        \caption{Traditional Compensation}
        \label{Traditional Method Detection}
    \end{minipage}
    \hfill
    \begin{minipage}[t]{0.30\columnwidth}
        \centering
        \includegraphics[width=\textwidth]{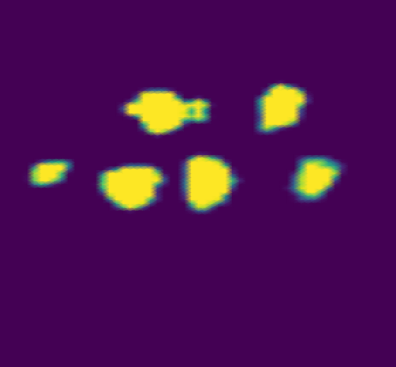}
        \caption{LLM-based Compensation}
        \label{LLM-Assisted Detection}
    \end{minipage}
    \hfill
    \begin{minipage}[t]{0.30\columnwidth}
        \centering
        \includegraphics[width=\textwidth]{Figures/Raw_data_to_LLM.png}
        \caption{Default Order}
        \label{Noise Removal First}
    \end{minipage}
    \vfill
    \begin{minipage}[t]{0.30\columnwidth}
        \centering
        \includegraphics[width=\textwidth]{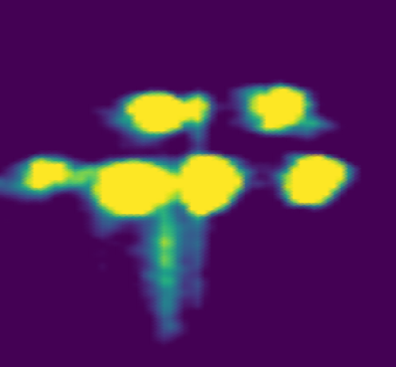}
        \caption{Swapped Order}
        \label{Power Compensation First}
    \end{minipage}
    \hfill
    \begin{minipage}[t]{0.30\columnwidth}
        \centering
        \includegraphics[width=\textwidth]{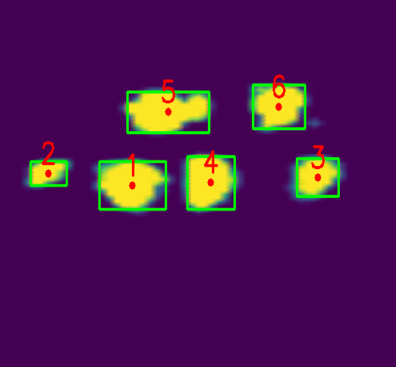}
        \caption{Detection with Data Enhancement}
        \label{Submit data with data enhancement}
    \end{minipage}
    \hfill
    \begin{minipage}[t]{0.30\columnwidth}
        \centering
        \includegraphics[width=\textwidth]{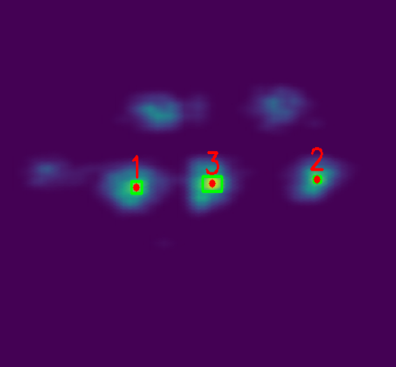}
        \caption{Detection with Raw Data Input}
        \label{Submit data without data enhancement}
    \end{minipage}
\end{figure}

\vspace{-2.5mm}
\section{Evaluation}
\label{sec: evaluation}
We evaluate the performance of our system in two key metrics: \textbf{Counting Accuracy} and \textbf{Grid Average Mean Error (GAME)}. To ensure comprehensive evaluation across multiple dimensions, we conduct the tests in real-world setups along with an ablation study to verify the effectiveness of each module.
     
\subsection{Real-world Experiments}
\begin{figure}[h]
    \centering
    \includegraphics[width=\linewidth]{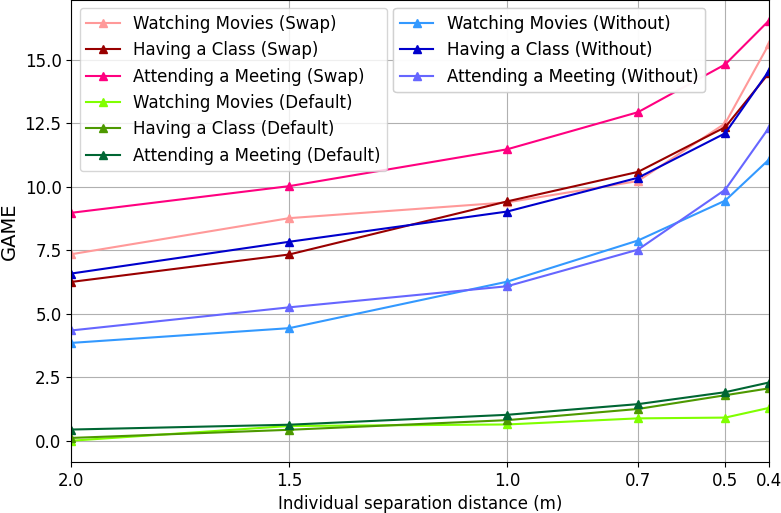}
    \caption{Ablation Study GAME as Metric}
    \label{Ablation Study GAME as Metric}
\end{figure}

\begin{figure}[h]
    \centering
    \includegraphics[width=\linewidth]{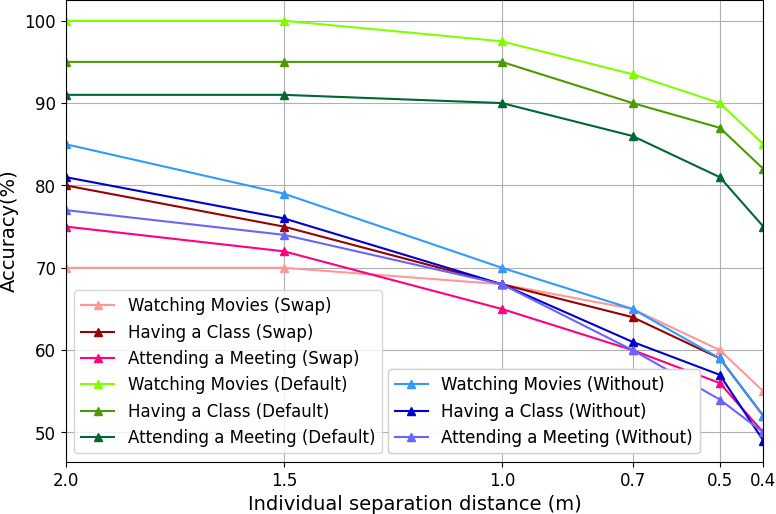}
    \caption{Ablation Study Accuracy as Metric}
    \label{Ablation Study Accuracy as Metric}
\end{figure}

We evaluate the system in various real-world scenarios, including 'Watching Movies', 'Having a Class', and 'Attending a Meeting'. Volunteers are seated at distances from 0.4m to 2.0m, sitting naturally. During each 15-minute session, samples are collected every 10 seconds to assess counting accuracy and Grid Average Mean Error (GAME). GAME measures the average error across individual grids, reflecting the discrepancy between the original and approximated grids. It is calculated as:

\[GAME=\frac{1}{N}\sum^N_{i=1}\sum^{4^L}_{l=1}|C^{pred}_{i}(l) - C^{gt}_{i}(l)|\]
where \(L\) represents represents the level of division. For each increase in \(L\), the image is divided into \(4^L\)  grid cells, and \(l\) represents the \(lth\) grid patch. \\
% GAME is more robust to localization errors in density estimation.
We illustrate the experiment results in Figure\ref{Ablation Study GAME as Metric} and Figure \ref{Ablation Study Accuracy as Metric}, noted as `Default'.
\vspace{-5.0mm}
\vspace{-4mm}
\vspace{-5mm}
\subsection{Ablation Study}
% \noindent\textbf{Swapping Noisy Removal and Data Compensation}
% \noindent\textbf{Submitting data with/without data enhancement to the LLM}\\

%Fig. 11 和 Fig. 12是什么意思？ 这里得对比data enhancement前后的差别
% Have corrected -boyan
% Figure \ref{Submit Raw data to LLM} displays clearer and more distinct heat spots, making it easier to count and analyze individual areas of interest. In contrast, Figure \ref{Submit Heatmap to LLM} processed using the heatmap input, shows more diffuse and less defined heat spots, complicating the task of accurate counting and analysis.

We first verify the effectiveness of the data enhancement order by swapping the steps of noise removal and dynamic power compensation: The detection performance in Figrue \ref{Ablation Study GAME as Metric} and Figure \ref{Ablation Study Accuracy as Metric} shows error occurred after the swapping while the comparison between the results visualized in Figure \ref{Noise Removal First} and Figure \ref{Power Compensation First} reveals that significant noise arises when the order is reversed. This occurs because the noise isn't removed before the power compensation process, causing the noise to be amplified along with the signal during the compensation process.

We then verify the necessity of the whole data enhancement module by performing the detection directly without one. Figure \ref{Submit data with data enhancement} and Figure \ref{Submit data without data enhancement} shows the visualized detection results with/without the data enhancement, then we compare the detection performance in Figrue \ref{Ablation Study GAME as Metric} and Figure \ref{Ablation Study Accuracy as Metric}, which reveals a considerable performance drop, demonstrating the effectiveness of this module.

% Through comparison between the two images reveals that the sequence of processing steps greatly impacts  clarity and quality. In Figure \ref{Noise Removal First}, the noise was effectively minimized, leading to a cleaner and more focused enhancement of high-intensity areas. On the other hand, in Figure \ref{Power Compensation First}, there is compensation noise alongside the signal, causing a less distinct image with more noise artifacts. 

\bibliographystyle{IEEEbib}
\bibliography{refs}

\end{document}